\documentclass[10pt,twocolumn,letterpaper]{article}

\usepackage{iccv}
\usepackage{times}
\usepackage{epsfig}
\usepackage{graphicx}
\usepackage{amsmath,amsthm}
\usepackage{amssymb}
\usepackage{comment}
\usepackage[font=small]{caption}
\usepackage[labelformat=simple]{subcaption}

\usepackage{multirow}
\usepackage[table,x11names]{xcolor}

\usepackage[pagebackref=true,breaklinks=true,letterpaper=true,colorlinks,bookmarks=false]{hyperref}

\iccvfinalcopy 

\def\iccvPaperID{3} 

\ificcvfinal\fi
\begin{document}

\title{Geo-Aware Networks for Fine-Grained Recognition}

\author{Grace Chu \qquad Brian Potetz \qquad  Weijun Wang \qquad Andrew Howard \and Yang Song \qquad Fernando Brucher \qquad  Thomas Leung \qquad Hartwig Adam\\
Google Research\\
\{cxy, potetz, weijunw, howarda, yangsong, fbrucher, leungt, hadam\}@google.com
}

\maketitle
\thispagestyle{empty}

\begin{abstract}

Fine-grained recognition distinguishes among categories with subtle visual differences. In order to differentiate between these challenging visual categories, it is helpful to leverage additional information. Geolocation is a rich source of additional information that can be used to improve fine-grained classification accuracy, but has been understudied. Our contributions to this field are twofold. First, to the best of our knowledge, this is the first paper which systematically examined various ways of incorporating geolocation information into fine-grained image classification through the use of geolocation priors, post-processing or feature modulation. Secondly, to overcome the situation where no fine-grained dataset has complete geolocation information, we release\footnote{https://github.com/visipedia/fg\_geo} two fine-grained datasets with geolocation by providing complementary information to existing popular datasets - iNaturalist and YFCC100M. By leveraging geolocation information we improve top-1 accuracy in iNaturalist from 70.1\% to 79.0\% for a  strong baseline image-only model. Comparing several models, we found that best performance was achieved by a post-processing model that consumed the output of the image-only baseline alongside geolocation. However, for a resource-constrained model (MobileNetV2), performance was better with a feature modulation model that trains jointly over pixels and geolocation: accuracy increased from 59.6\% to 72.2\%. Our work makes a strong case for incorporating geolocation information in fine-grained recognition models for both server and on-device.

\end{abstract}

\vspace{-3mm}
\section{Introduction}

Fine-grained recognition helps people distinguish subordinate categories of an object, e.g. recognizing the species of cats, dogs, flowers, etc. \cite{inat2017_comp}. It is a challenging problem as the visual difference among fine-grained categories is subtle \cite{fg_subtle_diff}. Moreover, images are often photographed at angles that fail to capture the subtle difference. To overcome these difficulties, researchers have been using various forms of complementary information besides image pixels to help with fine-grained recognition, such as attributes, poses, and text \cite{fg_attributes_airplane,fg_pose_norm,fg_text}.

\begin{figure}
    \centering
    \includegraphics[width=70mm]{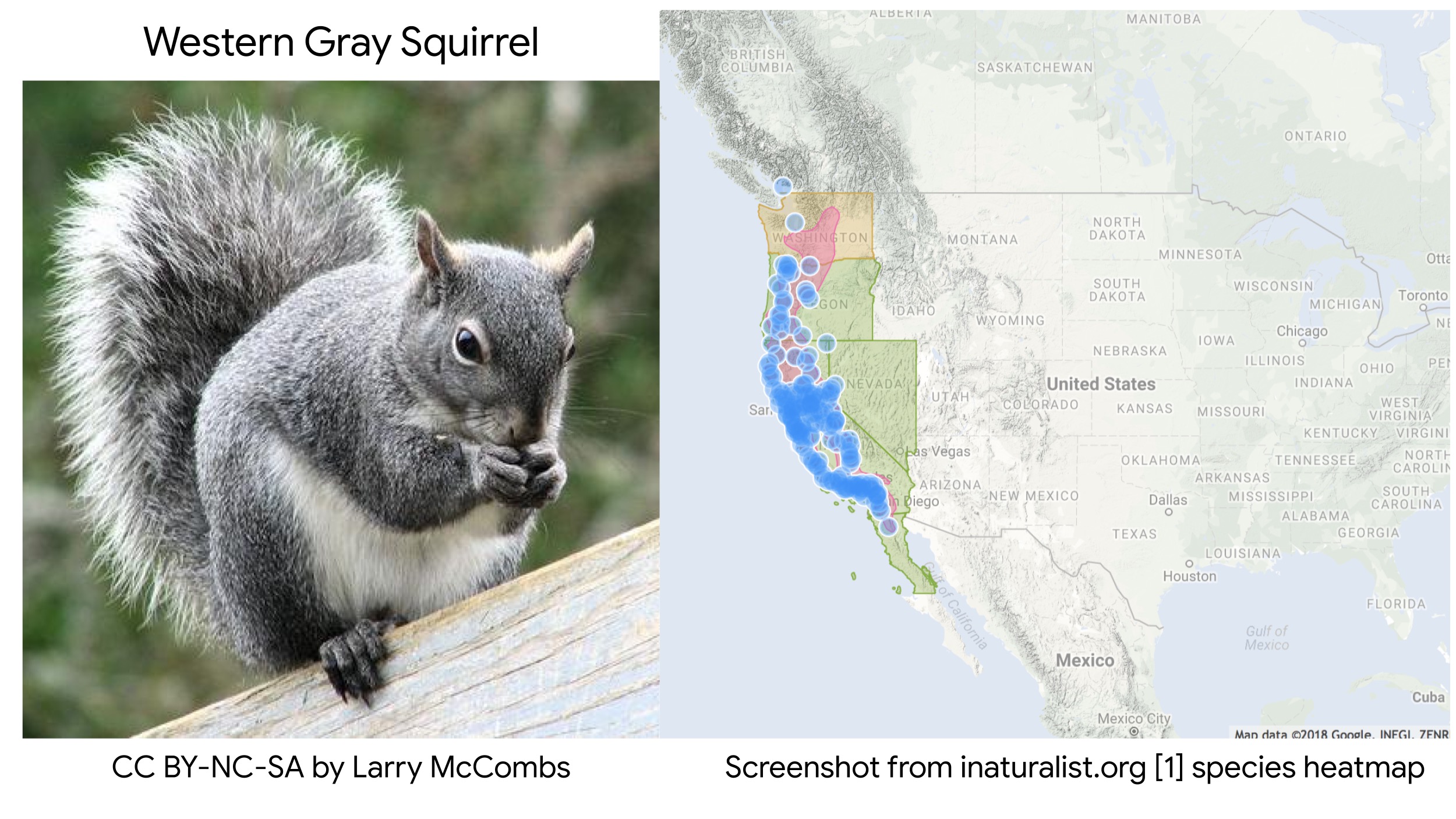}
    \caption{Western gray squirrel and its habitat heatmap.}
    \label{fig:intro_example}
    \vspace{-3mm}
\end{figure}

Geolocation has already been proven to be useful to distinguish coarse-grained classes, such as bridges and monuments \cite{yfcc100m_geo,geo_birdsnap}. But the benefits of purely using raw latitude and longitude (lat/lon) was small, while the bulk of the improvements came from integrating extra features, like Instagram hashtags associated with different geographical regions \cite{yfcc100m_geo}. For fine-grained recognition, on the other hand, geolocation may play a much bigger role because the geolocation distribution of a fine-grained object, like western gray squirrel in Figure \ref{fig:intro_example}, is generally more concentrated than that of a coarse-grained object, like dog. Thus, geolocation may be more effective to disambiguate species than general objects. Also, visually distinguishing fine-grained classes is generally harder than coarse-grained classes, which gives more room to improve via other orthogonal signals like geolocation.

In this paper, we systematically examine the effectiveness of using geolocation on fine-grained recognition problems and show that by only using raw lat/lon, we can achieve significant improvements upon state of the art image-only models \cite{inat2017_comp}. The improvement of using raw lat/lon on fine-grained dataset iNaturalist \cite{inat2017_data} (8.9\%) is even bigger than that using 6 lat/lon derived extra features on coarse-grained dataset YFCC100M-GEO \cite{yfcc100m_geo} (7\%). 

Specifically, we first examined an intuitive way of using geolocation priors where we discussed both the Bayesian approach and a whitelist-based method. Then, we examined a post-processing method where a geolocation network is combined with a pre-trained and frozen image network at the logits layer. Significant improvement has been observed using this model. Finally, we examine geolocation's impact on image feature learning through a feature modulation approach, which significantly outperforms other methods for the case of mobile resource constrained models.

In order to demonstrate the effectiveness of our geo-aware models, we introduce two fine-grained datasets with geolocation information. Both are based on existing datasets, but with additional fine-grained labels or added geolocation information.

The rest of the paper is organized as follows. Section \ref{sec:related_works} gives an overview of related works. Section \ref{sec:geo_networks} presents the three geo-aware networks we examine in this paper. In Section \ref{sec:geo_data}, two fine-grained datasets with geolocation are introduced. Then, experimental results are demonstrated in Section \ref{sec:experimental_results}. Section \ref{sec:conclude} concludes the paper.


\section{Related Works}
\label{sec:related_works}

\textbf{Fine-grained recognition} differs from general visual recognition mainly due to the following two aspects: different fine-grained categories usually have little visual difference that only domain experts can tell; rare subordinate objects are observed less while commonly seen ones dominant the fine-grained dataset. This leads to a long-tail label frequency distribution in such problems \cite{fg_long_tail}. Therefore, although the advances of general convolutional neural networks (CNN) \cite{inception_v3, mobilenet_v2} can lead to progress in fine-grained recognition, there is still more research needed in this area.

To deal with the subtle visual difference of fine-grained recognition, researchers have tried various directions. Among different model architectures, bilinear CNN has been proven to be effective through learning localized feature interactions \cite{cuiyin_bilinear}. Attention networks have also been used to locate the subtle difference between fine-grained labels \cite{fg_attention_ma_cnn, fg_attention_ra_cnn}. Besides visual information, researchers have been using additional information such as pose \cite{fg_pose_norm}, attributes \cite{fg_attributes_airplane} and text description \cite{fg_text}. Data augmentation and transfer learning have also been studied \cite{fg_web_augment, inat2017_comp}.

\textbf{Geolocation} has been widely used for coarse-grained classifications. Tang \textit{et al.} \cite{yfcc100m_geo} used 6 geolocation related features and concatenated them with the image model output before the softmax to improve classification accuracy on classes like snow, monument and wave. One of the 6 geolocation related features in this work is latitude and longitude, while other features incorporated extra information, such as geographic maps and hashtags in Instagram. To solve similar problems, Liao \textit{et al.} \cite{geo_neighbor_labels} approached it by finding neighbor images taken near the target image, and then used the tag distribution of neighbor images as a feature to feed into support vector machine (SVM) classifiers. Geolocation has also been used for scene understanding \cite{geo_region} and place identification \cite{geo_place_type}.

There are, however, only few works in fine-grained recognition that have tried using geolocation to improve accuracy. Berg \textit{et al.} in \cite{geo_birdsnap} made a simulated geolocation fine-grained dataset by combining an image-only dataset and a geo only dataset. Then, Bayesian based geolocation priors were used to improve the classification accuracy. Some participants of PlantCLEF2016 competition \cite{plantclef2016} tried using geolocation information. The competition contains plant species in and around France where only a minority of the images contain geolocation. A few non-neural network based methods were tried, but with no obvious improvements \cite{geo_dist_france, plantclef_meta}.


\section{Geo-Aware Networks}
\label{sec:geo_networks}

In this section, we study three methods to integrate geolocation with image feature based fine-grained models.

\subsection{Geolocation Priors}
\label{sec:geo_priors}

As discussed in the introduction, animal or plant species are distributed on the earth with some geographical traits. Assuming the data samples containing geographical information are observed independently in both the training and test datasets, we can extract the geolocation based distribution from the training data. There are two intuitive ways of utilizing this distribution without additional model training or any change to the image-only classifier.

\textbf{Bayesian Priors:}
From the Bayesian inference point of view, given image observation $I$ without additional information, traditional fine-grained recognition can be viewed as a Maximum Likelihood Estimation (MLE).
\begin{equation}
\vspace{-2mm}
\hat{L}_{MLE}(I) = \arg\max_{L} f(I | L),
\end{equation}
\vspace{-1mm}
where $L$ denotes the image label and $f(I|L)$ denotes the likelihood function of an observation given the label .

Now assume that a prior distribution $P(L | G)$ over the fine-grained labels exists and follows some geographical traits, where $G$ denotes the geolocation of the examined image. Then, it allows us to make a Maximum A Posteriori (MAP) estimation:
\begin{equation}
\vspace{-2mm}
\hat{L}_{MAP}(I,G) = \arg\max_{L} f(I | L)P(L|G).
\end{equation}
\vspace{-1mm}

\textbf{Label Whitelisting:}
A different way of utilizing the geographical information is to restrict the inference result by a geo-restricted whitelist. For example, if an image is taken in a certain city or a zoo, then only labels that have been observed in that city or zoo will be presented to the user. The geo-restricted whitelist works as a gating function which restricts output labels to be one in the whitelist of labels that have data observations within a geo-restricted radius $\theta$:
\begin{equation}
\vspace{-2mm}
\hat{L}_{MAP}(I,G) = \arg\max_{L} f(I | L)\textbf{1}_{\theta} (L , G),
\end{equation}
\vspace{-1mm}
where $\textbf{1}_{\theta}(L,G)$ is an indicator function, which equals to one when $L$ has observsations within geo-restricted radius $\theta$ of G, zero otherwise.


\subsection{Post-Processing Models}
\label{sec:post_processing_models}

We consider a post-processing model to be any model that does not touch pixels, but instead consumes one or more image classifiers or embeddings. Here, we trained a post-processing model that consumes the output of the baseline image classifier together with geolocation coordinates.

The model evaluated below accepts geolocation in its simplest form: a vector of length two containing latitude and longitude, normalized to range $[-1, 1)$. We also experimented with Earth-center-fixed rectangular coordinates and multi-scale one-hot S2 representations \cite{s2_cells}. These made little difference in performance.

\begin{figure}
    \centering
    \includegraphics[width=85mm]{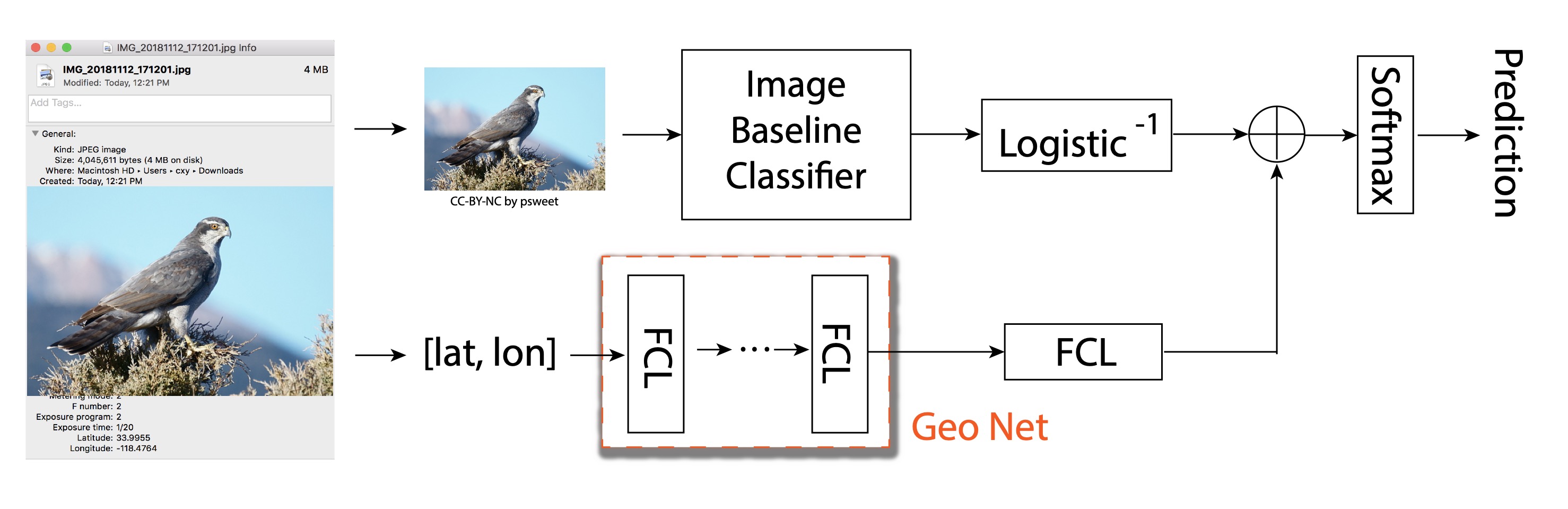}
    \caption{Network architecture for post-processing models. Logistic$^{-1}$ is the inverse function of logistic function. ``FCL" denotes fully connected layer. The last FCL outside geo net box is the logits layer.}
    \label{fig:auto_geo}
     \vspace{-3mm}
\end{figure}

Geolocation is then processed by three fully connected layers of sizes 256, 128, and 128, followed by a layer of logits with size equal to the output label map. These are then added to the logits of the image classifier, or $\sigma^{-1}(output)$, where $\sigma(x) = 1/(1+e^{-x})$. Figure \ref{fig:auto_geo} shows the diagram of the architecture. In this late-fusion architecture, no units jointly encode appearance and location. We also experimented with models where the output of the image classifier or an image embedding was concatenated with one of the fully-connected layers in the geolocation network. In these models, units jointly encode appearance and location. However, adding these visual inputs does not affect performance of the post-processing model. This may suggest that appearance and location are not tightly interdependent.

During training, the weights of the baseline image classifier are fixed, and gradients are not pushed back through the image classifier. One disadvantage of this is that the image classifier may waste effort attempting to distinguish two visually-similar labels that could have been easily distinguished by geolocation alone.

Post-processing models offer some practical advantages over jointly training over pixels and geolocation. Learning rates, hyperparameters, and loss functions are decoupled between the two models, and can be tuned separately. Similarly, the selection and balancing of training data can be performed independently for image models versus geolocation models. If labeled training images without geolocation are available, they can be used to train the image classifier but omitted when training the post-processing model. If label noise is correlated with appearance but not with location (e.g. mustang cars mixed in with horses), then the post-processing model may benefit from the inclusion of noisier training data sources that harm image classifier performance.

Another feature of the post-processing model is that the geolocation network only needs to learn the residual between the baseline image classifier output and the ground-truth. If the baseline image classifier already classifies a label perfectly, then no geolocation model will be learned for that label, since none is needed. Thus, the post-processing model minimizes its reliance on geolocation cues: it relies on geolocation only in proportion to how much it improves an image classifier that was previously trained to maximize performance.

Adding the logits of the geolocation and image networks has some theoretical basis. Suppose appearance and location were conditionally independent of each other given the ground-truth label (so that the appearance of a label does not change depending on its location). Then $P(L,G|I) = P(L|I)P(G|L)$, where $I$ is the image, $G$ is the geolocation, and $L$ is the ground-truth label. For convenience, define the likelihood ratio $R = P(G|L)/P(G|\bar{L})$, where $\bar{L}$ denotes the condition that label $L$ is false. Then:
\vspace{-1mm}
\begin{align*}
  P(L|I,G) &= \frac{P(L,G|I)}{P(L,G|I) + P(\bar{L},G|I)}  \\
           &= \frac{P(L|I)P(G|L)}{P(L|I)P(G|L) + P(\bar{L}|I)P(G|\bar{L})}  \\
           &= \frac{P(L|I)R}{P(L|I)(R-1) + 1}  \\
           &= \left(1 + \frac{1}{RP(L|I)} - \frac{1}{R} \right)^{-1}  \\
           &= \sigma \left( -log \left(\frac{1}{RP(L|I)} - \frac{1}{R} \right) \right)   \\
           &= \sigma \left( -log \left(\frac{1}{P(L|I)} - 1 \right) + log(R) \right)   \\
           &= \sigma \left( \sigma^{-1}(P(L|I)) + log(R) \right).
\end{align*}

\vspace{-1mm}
Thus, if conditional independence holds, then the post-processing network would be optimal in the sense that it outputs the exact posterior probability $P(L|I,G)$ if the output of image classifier equals $P(L|I)$ and the geolocation logits activation equals $log(R)$. Conditional independence is a sufficient condition for the model to behave optimally for some set of weights, but not a necessary condition. For example, suppose geolocation could sometimes be estimated from the background of the image. In this scenario, models using Bayesian priors would double-count location evidence, adjusting scores based on location even though the image classifier already factored it in. In contrast, since the post-processing model trains the geolocation network based on the residual between the baseline image classifier output and the ground-truth, no double-counting occurs; the learned geolocation model is only as strong as geolocation evidence not already captured by the baseline image classifier.


\subsection{Feature Modulation Models}
\label{sec:geo_feature_modulation}

To examine whether geolocation can have a deeper effect on image feature learning, we built networks with geolocation information integrated into the image features.

\begin{figure}
    \centering
    \includegraphics[width=85mm]{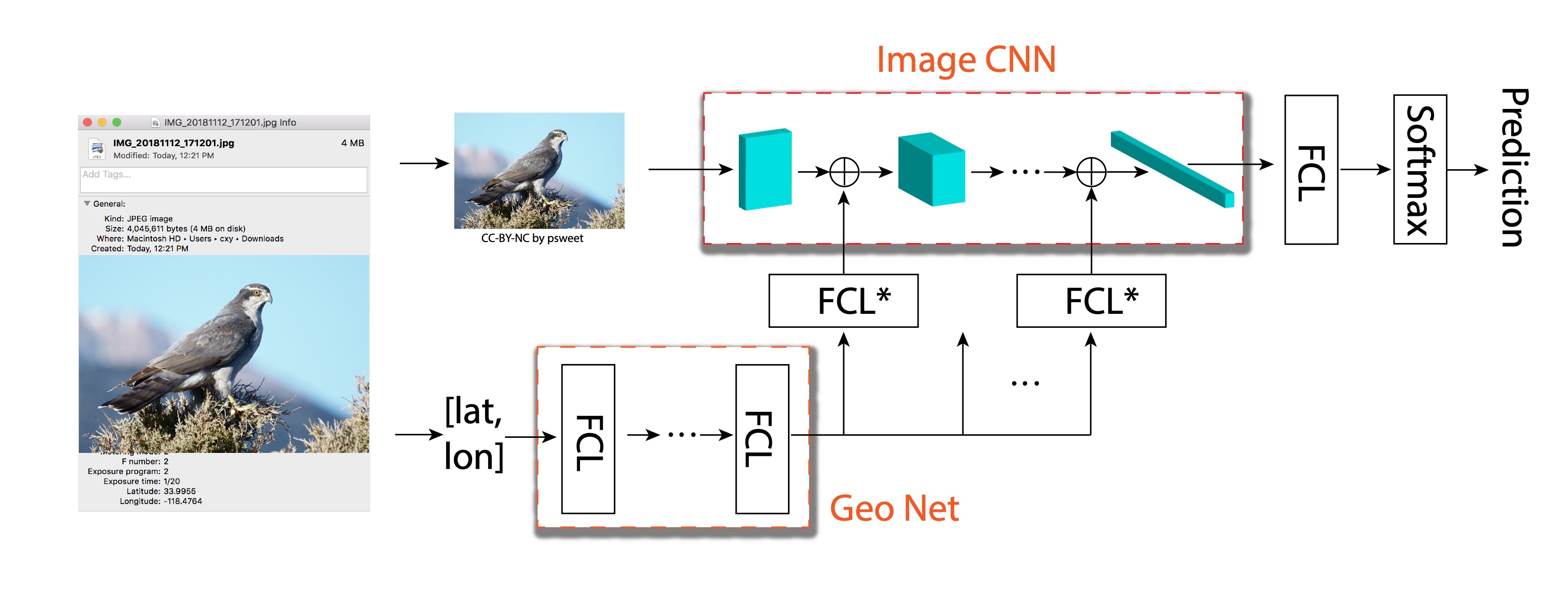}
    \caption{Network architecture of using geolocation to affect image features. FCL* represents FCL without activation, and has a reshape operation afterwards to match the feature dimension it adds to.}
    \label{fig:geo_features}
     \vspace{-3mm}
\end{figure}

Similar to post-processing model, we use addition to modulate image features via geolocation features. As shown in Figure \ref{fig:geo_features}, latitude and longitude first go through a set of fully connected layers. Then, depending on the shape of each image feature, the output of geolocation network goes through different sized fully connected layers (without activation) to be reshaped before adding to the image feature. Mathematically,
\vspace{-2mm}
\begin{equation}
F_{post-act}^* = F_{post-act} + \beta,
\vspace{-1mm}
\end{equation}
where $F$ and $F^*$ are image features before and after modulation. Subscript ``post-act" indicates that the features are modulated after activation. $\beta$ are reshaped geolocation features.

Not all image features from each layer are modulated by geolocation features. Lowest level image features are general features specifying lines or edges of the object, which conveys little information about species level distinction. Thus, we only modulate middle and higher image features instead of lower ones. We also experimented with models that modulated all image features, but didn't get better results.

Perez \textit{et al.} \cite{film} introduced a generic feature modulation called FiLM. Specifically, they modulated image features by both multiplication and addition as follows:
\vspace{-2mm}
\begin{equation}
F_{pre-act}^* = \gamma * F_{pre-act} + \beta,
\vspace{-1mm}
\end{equation}
where subscript ``pre-act" indicates that the features are modulated before activation. $\gamma$ and $\beta$ are modulation features. In Section \ref{sec:experiments_feature_modulation}, we will show that, for geo-aware networks, only using addition is the best way to modulate image features.



\section{Fine-Grained Datasets with Geolocation}
\label{sec:geo_data}

One challenge of using geolocation in fine-grained recognition is the lack of fine-grained datasets with geolocation information. To the best of our knowledge, there are only two fine-grained datasets that have been used in geolocation related research in this field \cite{geo_birdsnap, plantclef2016}. In \cite{geo_birdsnap}, the authors simulated a geolocation fine-grained dataset by matching images from an image-only dataset with random observations from a geolocation only dataset with the same ground-truth label. The dataset for one of the ImageCLEF/LifeCLEF competitions \cite{imageclef}, PlantCLEF2016 \cite{plantclef2016}, contains partial geolocation information (less than half of the data) and is restricted to only plants from France. 

In this section, we will introduce two fine-grained datasets with geolocation, one for both training and evaluation; the other for evaluation only. Both datasets contain genuine (not simulated) and worldwide geolocation.

\subsection{iNaturalist Dataset with Geolocation}
\label{sec:inat}


We introduce the iNaturalist fine-grained dataset with geolocation based on the data from iNaturalist challenge at FGVC (fine-grained visual categorization) 2017. The challenge data, without geolocation, was published in \cite{inat2017_data} and available in the challenge page \cite{inat2017_challenge}. The state-of-the-art classification results based on this dataset were presented in \cite{inat2017_comp}. This dataset contains 5089 fine-grained labels. To be comparable with existing results, we used the same train/test split as in \cite{inat2017_comp}, where 665,473 images are in training and 9,697 images are in test.

To obtain geolocation information of above dataset, we first map image keys in \cite{inat2017_challenge} to observation ids. Then, we utilize the iNaturalist observation data from Global Biodiversity Information Facility (GBIF) \cite{gbif_inat} which contains observation ids and geolocation data. From the path of image keys to observation ids to geolocation, we can find the corresponding geolocation information for existing iNaturalist challenge images.

During the mapping process, there are $\sim$4\% images that couldn't find corresponding geolocation information, due to either missing observation ids in \cite{inat2017_challenge} or missing geolocation in the GBIF observation data. The final geolocated dataset contains 645,424 images in training and 9,394 images in test. Figure \ref{fig:geo_dist_a} shows the heatmap of the geolocation distribution of the obtained dataset, including both training and test data. This indicates the worldwide distribution of our iNaturalist based fine-grained geolocation dataset.


\subsection{YFCC100M Fine-Grained Evaluation Dataset with Geolocation}

\begin{figure}
    \centering
    \begin{subfigure}[t]{0.48\columnwidth}
      \includegraphics[width=\columnwidth,keepaspectratio]{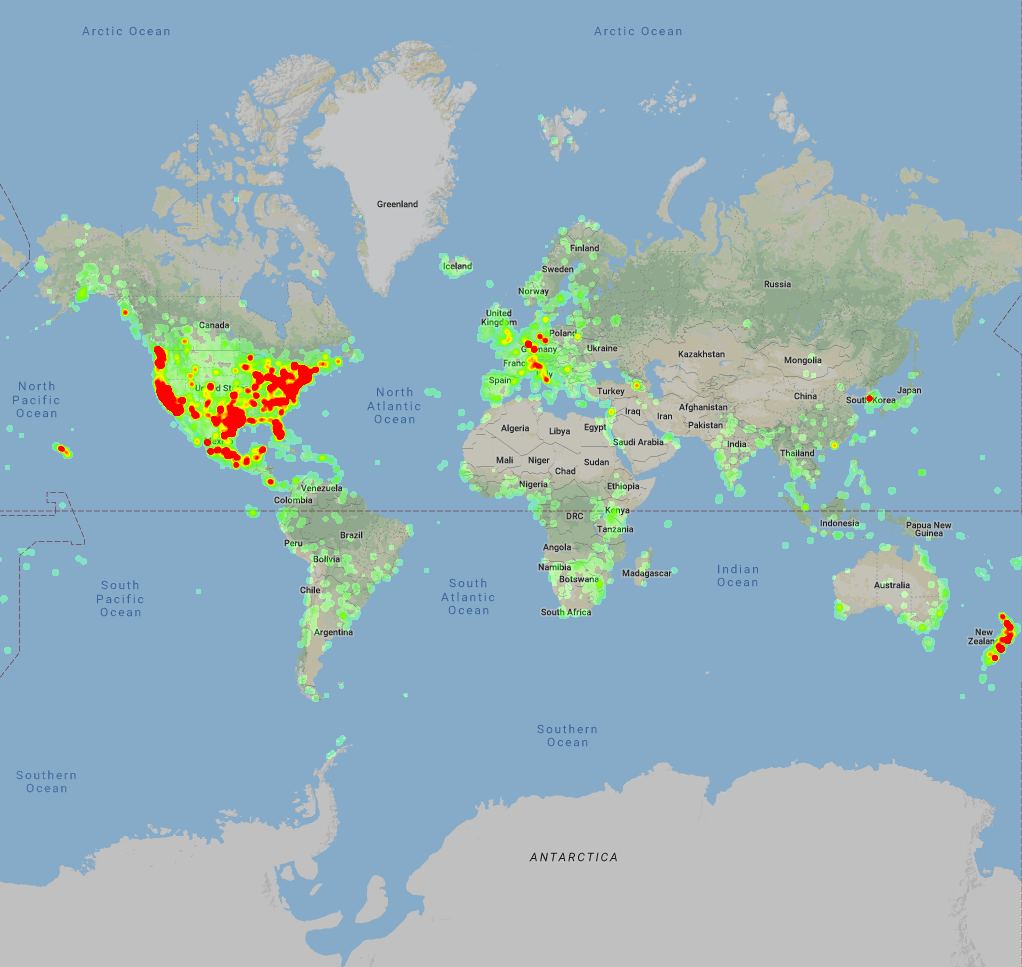}
      \caption{}
       \vspace{-3mm}
      \label{fig:geo_dist_a}
    \end{subfigure}
    \hfill
    \begin{subfigure}[t]{0.48\columnwidth}
      \includegraphics[width=\columnwidth,keepaspectratio]{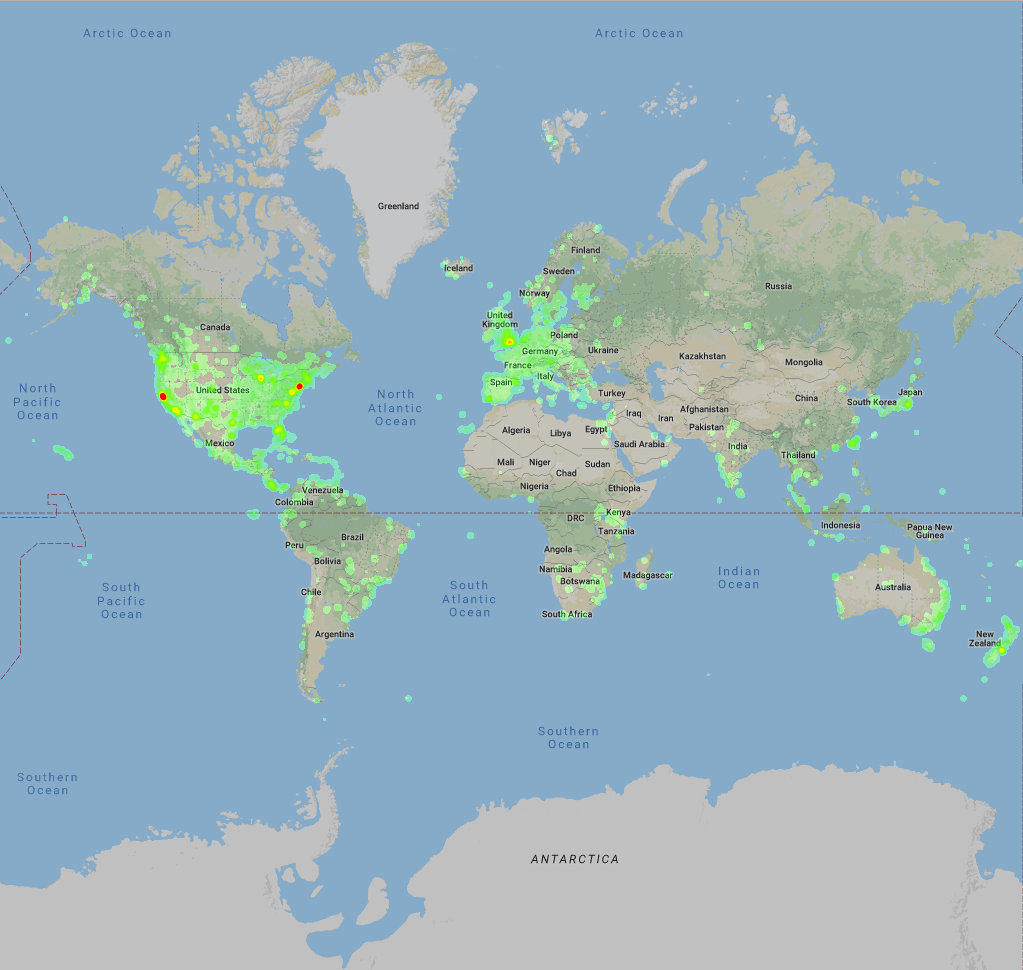}
      \caption{}
      \label{fig:geo_dist_b}
       \vspace{-3mm}
    \end{subfigure}
    \caption{Geolocation distribution of \subref{fig:geo_dist_a} iNaturalist dataset, and \subref{fig:geo_dist_b} YFCC100M fine-grained evaluation dataset.}
    \label{fig:geo_dist}
     \vspace{-3mm}
\end{figure}

The YFCC100M dataset consists of 100 million Flickr images and videos with creative commons licences \cite{Thomee:2016}. For each image, we identified Flickr tags or image titles that contained labels corresponding to one of the 5089 fine-grained plant and animal species labels from the iNaturalist dataset in Section \ref{sec:inat}. Since iNaturalist labels are all species-level, images with multiple labels were omitted. 1,362,447 geolocated images had a single matching label. iNaturalist labels in the YFCC100M dataset are highly skewed towards popular species like domestic animals, cut flowers, and zoo animals. To mitigate the impact of highly common labels, we limited our evaluation to at most 10 examples per label. Of 4721 labels represented in YFCC100M, 3553 labels had at least 10 examples. 36,146 labeled geolocated images were used in total. The distribution heatmap of geolocations of these images are shown in Figure \ref{fig:geo_dist_b}, which has similar coverage as iNaturalist geolocation dataset.

\section{Experimental Results}
\label{sec:experimental_results}

In this section, we present experimental results of the examined geo-aware networks. To show the effectiveness of using geolocation, we compare geo-aware networks with the state-of-the-art image-only model presented in \cite{inat2017_comp}. Specifically, we take the Inception V3 with 299x299 input size as the image baseline classifier. From this initial checkpoint, we train or calculate results for our geo-aware networks based on the iNaturalist dataset with geolocation. We evaluate each model over the independent YFCC100M dataset to show the generalization of the geo-aware networks. 
Finally, we show how performance is affected in a mobile on-device setting, using a MobileNetV2 baseline.

\subsection{Geolocation Priors}

As discussed in Section \ref{sec:geo_priors}, we assume that the geolocation priors follow certain geographical traits, therefore the prior distribution will differ as geolocation changes. To use the geolocation based prior distribution for inference, we treat the geolocation of each testing data sample as a reference point. For each reference point, all training data points within a certain radius from this referenced geolocation are counted with equal weights and a histogram of class labels is calculated. After this, we either use the histogram as a whitelist of labels or normalize it to get a prior probability distribution.

We empirically pick the best radius for the best baseline accuracy numbers using geolocation priors. We picked a few radius in the range sweeping from 50 miles to 5000 miles and found 100 miles to be the golden number for iNaturalist dataset. We also found that using geolocation based Bayesian prior produces worse results on iNaturalist, which is likely due to fact that the geolocation based prior distributions in the test set are more uniform and mismatch the ones estimated from the training set. Using a label whitelist mitigates the disparity between the prior distribution on the training set and the one on the test set, which gives better results. More quantitative results are given in Table \ref{tb:geo_prior}.

\vspace{-1mm}
\begin{table}[ht]
\caption{Top-1 accuracies using geolocation based Bayesian priors and whitelist with different radius (miles) at each test location on iNaturalist dataset; the image-only baseline model gives $70.1\%$ \cite{inat2017_comp}}
\vspace{-2mm}
\centering
  \begin{tabular}{ l | c | c | c | c }
    \hline
      & 50 & 100 & 500 & 1000  \\
      \hline \hline
    Bayesian Priors & 68.5\% & 69.4\% & 67.8\% & 66.5\% \\ 
    Whitelisting & 71.3\% & \textbf{72.6}\% & 72.3\% & 71.8\% \\ 
    \hline
  \end{tabular}
  \label{tb:geo_prior}
  \vspace{-2mm}
\end{table}


\subsection{Post-Processing Models}

The post-processing models were trained over the iNaturalist training partition at a learning rate of 0.02 without decay. It consumed the output of the Inception V3 model described in Section \ref{sec:experimental_results}, without touching pixels. Evaluated over the iNaturalist evaluation set, it achieved 79.0\% accuracy for the top label, an increase of 8.9\% over the baseline model.


\subsection{Feature Modulation Models}
\label{sec:experiments_feature_modulation}

Experimental setups for feature modulation model are as follows. Take Figure \ref{fig:geo_features} as the reference, FCL layers inside geolocation network have output sizes of 128 then 256. We use Inception V3 as the image CNN and apply feature modulations for all image features out of the Inception modules \cite{inception_v3}. The whole network, including image CNN and geo net, are trained together end to end, where only the image CNN part has parameters initialization copied from the image-only baseline model. We have used \mbox{RMSprop} optimizer with initial learning rate 0.0045, decaying every 4 epochs with decay rate 0.94.

The last bolded line in Table \ref{tb:modified_film} shows the Top-1 and Top-5 accuracies of the proposed feature modulation model. Comparing with the image-only model (first line), our proposed geo-aware network achieves 8.1\% increase on top-1 accuracy and 3.9\% increase on top-5 accuracy.

\vspace{-1mm}
\begin{table}[ht]
\caption{Top-k accuracies for different feature modulations. $F$ and $\mathcal R(F)$ denote the image feature before and after (ReLU) activation. $\mathcal R(\cdot)$ and $\mathcal S(\cdot)$ denote ReLU and Sigmoid activation function respectively. $\gamma$ and $\beta$ are geolocation features which are the outputs of two different geo networks followed by the reshape FCL* for each modulation layer.}
\vspace{-2mm}
\centering
  \begin{tabular}{ l | c  c }
    \hline
    & Top-1 & Top-5  \\
    \multirow{-2}{*}{Feature Modulation} & Accuracy & Accuracy \\ 
    \hline \hline
    None \cite{inat2017_comp} & 70.1\% & 89.4\% \\ 
    FiLM: $\mathcal R(\gamma \times F + \beta)$ \cite{film} & 72.5\% & 90.8\% \\ 
    \hline
    $\mathcal R(\gamma) \times \mathcal R(F) + \mathcal R(\beta)$ & 65.6\% & 87.1\% \\ 
    $\mathcal S(\gamma) \times \mathcal R(F) + \mathcal R(\beta)$ & 76.8\% & 93.1\% \\
    $\mathcal S(\gamma) \times \mathcal R(F)$ & 76.2\% & 92.9\% \\
    $\mathcal R(F) + \mathcal R(\beta)$ & 77.2\% & 93.1\% \\
    \boldmath $\mathcal R(F) + \beta$ & \textbf{78.2\%} & \textbf{93.3\%} \\
    \hline
  \end{tabular}
  \label{tb:modified_film}
  \vspace{-2mm}
\end{table}

The second line in Table \ref{tb:modified_film} shows results of using the general feature modulation scheme proposed in \cite{film}. The improvements over image-only model, 2.4\%, is less than one third of that obtained by our customized feature modulation model. In addition, we also tried other variations of the feature modulation, including modulating the feature before/after activation, using multiplication and/or addition as the modulation operation, and different activation functions on the modulator before combining with image features. We have listed some results in Table \ref{tb:modified_film}. However, none of them gives better results than our proposed method. Results demonstrate that addition is the preferred way to affect image features when using geolocation as the modulator.


\subsection{Comparison of Different Geo-Aware Networks}
\label{sec:experiments_compare}

\begin{figure*}
    \centering
    \includegraphics[width=175mm]{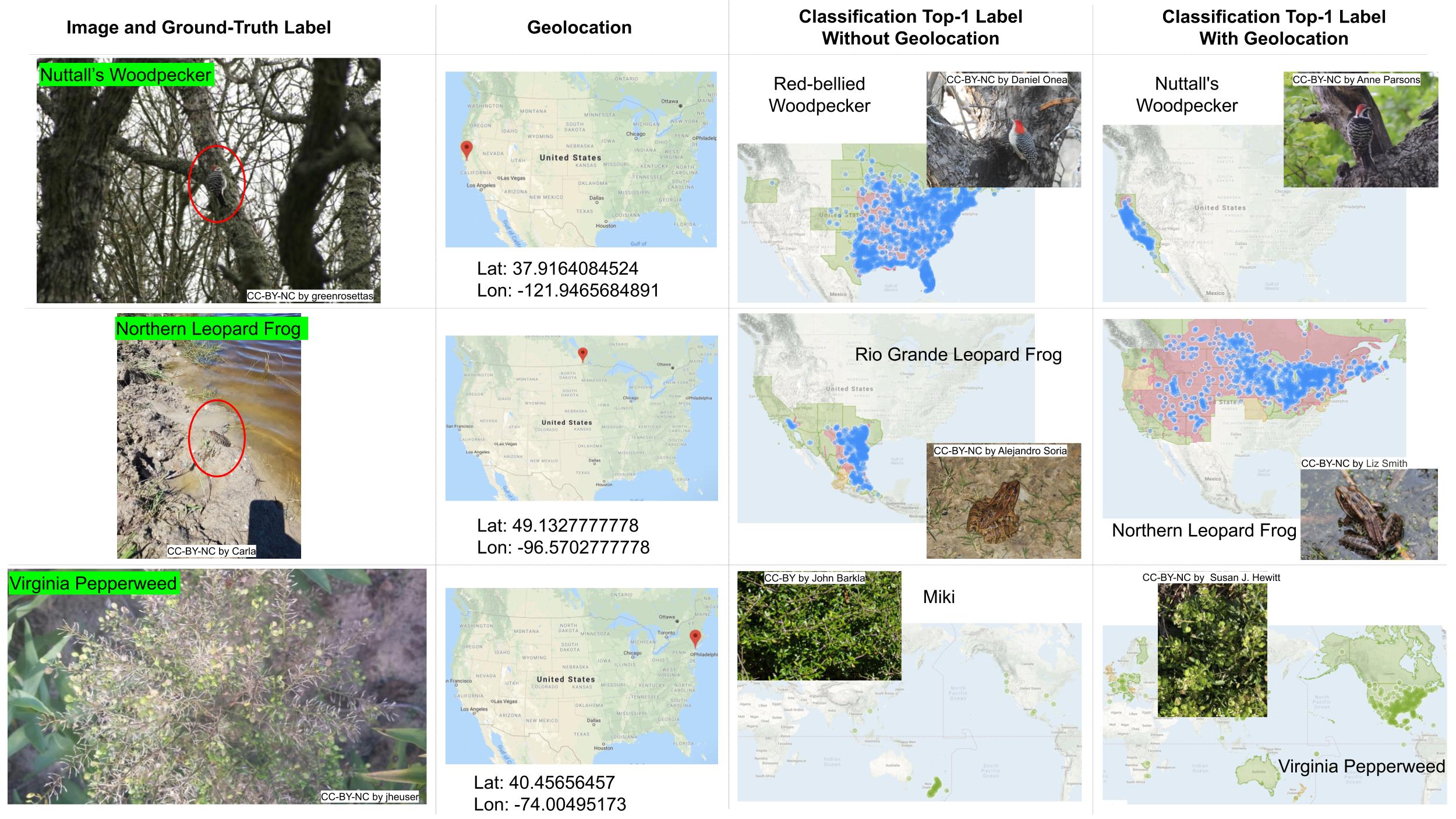}
    \caption{Examples where geo-aware networks corrected the prediction results using geolocation information. Distribution heatmaps are obtained by searching for the particular species/taxonomy in iNaturalist org \cite{inat_org}.}
    \label{fig:experimental_examples}
    \vspace{-3mm}
\end{figure*}

We summarize the best result from each network in Table \ref{tb:head_tail}. While all geo-aware networks achieve better results than image-only models, post-processing and feature modulation models give much better results than using geolocation priors. Among all models, post-processing model performs the best. The higher performance by post-processing model is informative because while feature modulation networks can capture arbitrary relationships $f(appearance, location)$, post-processing models are severely restricted in the relationships between appearance and location they can capture, expressing only $softmax(g(appearance) + h(location))$. This suggests that the dependencies between appearance and location that cannot be expressed by post-processing models may be rare in nature for fine-grained plants \& animals. 

\vspace{-1mm}
\begin{table}[ht]
\caption{Top-1 accuracies of different geo-aware networks, together with the head and tail results. Head and tail images are images whose labels have $\ge$ 100 images and $<$ 100 images in training set, respectively.}
\vspace{-2mm}
\centering
  \begin{tabular}{ l | c | c  c }
    \hline
    Geo-aware & Top-1 & Head: & Tail:  \\
      Model & Accu & $\ge$100 im & $<$100 im \\ 
      \hline \hline
    Image-Only & 70.1\% \cite{inat2017_comp} & 76.5\% & 66.2\% \\ \hline
    Whitelisting & 72.6\% & 77.2\% & 68.6\% \\ 
    Post-Process & \textbf{79.0\%} & 81.0\% & \textbf{77.2\%} \\ 
    Feature Modulate & 78.2\% & \textbf{81.1\%} & 75.6\% \\
    \hline
  \end{tabular}
  \label{tb:head_tail}
  \vspace{-2mm}
\end{table}

As fine-grained dataset usually has long tail distribution \cite{fg_long_tail}, we also show the results on head and tail images in Table \ref{tb:head_tail}. All geo-aware networks improve more on tail images than on head images. Specifically, the best post-processing model gives a 4.5\% increase on head images while having a 11\% increase on tail images, 2.4 times more improvement than on the head images. This implies that geolocation benefits more on lower baseline models which have more room to improve.

To better understand how geolocation improves the classification, we show some example images in Figure \ref{fig:experimental_examples} where geo-aware networks correct the wrong label given by the image-only model. Columns in this figure are, from left to right: image and its ground-truth label; geolocation of where this image was taken; top-1 label given by the image-only model, geo distribution heatmap of this label and a sample image of the same label randomly chosen from training dataset; top-1 label given by post-processing and feature modulation models (these two models give the same result for these examples), its corresponding geo distribution heatmap and a sample training image.

Take the first row as an example where the image-only model gives the wrong label - red-bellied woodpecker, while geo-aware models give the correct label - nuttall's woodpecker. By just looking at the sample images of these two labels/species, it is hard to visually distinguish them. However, they have completely different geolocation distributions which indicates their different habitats. Specifically, nuttall's woodpecker is only located on the west coast of America, while red-bellied woodpecker is mainly loacted in the center and the east coast of America. Therefore, when geo-aware models see that the image was taken on the west coast, they know that the bird in the image cannot be a red-bellied woodpecker whose habitat is in the center and east coast, and thus corrects the result.


\subsection{Results on Mobile Image Networks}
\label{sec:experiments_another_network}

While large models like Inception V3 give the best accuracy, their size and inference latency limits them to only running on server machines. However, there are cases where we need to run models on device due to connectivity, privacy, or speed concerns. In this subsection, we examine the performance of our geo-aware networks on a small on-device model: MobileNetV2.

We used the same settings as the ones used for Inception V3 in \cite{inat2017_comp} to train the MobileNetV2 image-only model on the iNaturalist dataset. Then, three geo-aware networks are calculated or trained based on this baseline image only classifier. For the feature modulation model, feature modulations are applied for all blocks with inverted bottlenecks. Table \ref{tb:diff_networks} shows the results on MobileNetV2 comparing to those on Inception V3.

\vspace{-1mm}
\begin{table}[ht]
\caption{Top-1 Accuracies of different geo-aware networks applied on different image baseline classifiers.}
\vspace{-2mm}
\centering
  \begin{tabular}{ l | c | c }
    \hline
    Geo-aware Model & Inception V3 & MobileNetV2  \\
      \hline \hline
    Image-Only & 70.1\% \cite{inat2017_comp} & 59.6\%  \\ \hline
    Whitelisting & 72.6\% & 62.1\% \\ 
    Post-Process & \textbf{79.0\%} & 70.7\% \\ 
    Feature Mod. & 78.2\% & \textbf{72.2\%} \\
    \hline
  \end{tabular}
  \label{tb:diff_networks}
  \vspace{-2mm}
\end{table}

Since the accuracy of the baseline is smaller, it has more room to improve. The best geo-aware network achieves 12.6\% top-1 accuracy increase over the image-only model, comparing with the 8.1\% increase for the larger model. Importantly, the best geo-aware network based on the MobileNetV2 model achieves even better performance than the image-only network based on Inception V3.

\vspace{-1mm}
\begin{table}[ht]
\caption{Top-1 accuracy of geo-aware networks, upon Inception V3 image baseline network, when evaluating on different evaluation data. FG denotes fine-grained.}
\vspace{-2mm}
\centering
  \begin{tabular}{ l | c | c  c  }
    \hline
    & Image & Post- & Feature \\
    \multirow{-2}{*}{Evaluation Dataset} & Only & Process & Mod. \\
    \hline \hline
    iNaturalist Eval & 70.1\% & 79.0\% & 78.2\% \\ 
    YFCC100M FG Eval & 54.6\% & 60.5\% & 58.7\%  \\
    \hline
  \end{tabular}
  \label{tb:diff_evals}
  \vspace{-2mm}
\end{table}

Unlike the results on Inception V3, feature modulation models outperform post-processing models on MobileNetV2 image baseline model. Recall that one disadvantage of the post-processing models is that the baseline image classifier it relies on must expend effort to distinguish visually-similar labels that can be easily disambiguated using geolocation. For a larger Inception model, this may be a small penalty. However, wasting capacity to visually distinguish, for example, American and European Magpies may be especially costly for a smaller on-device model.


\subsection{Results on YFCC100M Evaluation Data}
\label{sec:experiments_yfcc100m}

To demonstrate the generalization of the results in Section \ref{sec:experiments_compare}, the same models were also evaluated over our newly introduced YFCC100M fine-grained dataset. Results are shown in Table \ref{tb:diff_evals}. Post-processing model achieves 5.9\% gain over image-only model, while feature modulation model achieves 4.1\% gain. The improvements are smaller than those on iNaturalist evaluation set because the quality of this dataset is not as good as iNaturalist, which have been verified by domain experts. For example, some images in YFCC100M dataset contain animal sculptures instead of real animals, or the animal is mentioned in description and thus in the label but does not appear in the image.

\section{Conclusion}
\label{sec:conclude}

We have given a systematic overview of geo-aware networks for fine-grained recognition. To deal with the lack of fine-grained geolocation datasets, we introduced the iNaturalist and YFCC100M fine-grained geolocation datasets. Experimental results show that all geo-aware networks achieve significant improvements over image-only models. Specifically, the post-processing model performs best on large baseline models, while the feature modulation model performs best on small baseline models and even outperforms the large image-only model. Although experiments in this paper are mainly on animal and plant species recognition, we believe that the geo-aware networks examined in this paper are generally useful and can be easily extended for recognizing any location sensitive fine-grained categories, such as car's make/model and food.

\vspace{5mm}
\noindent \textbf{Acknowledgements} We would like to thank Yanan Qian, Fred Fung, Christine Kaeser-Chen, Professor Serge Belongie, Chenyang Zhang, Grant Van Horn and Oisin Mac Aodha for their help and useful discussions.

{\small
\bibliographystyle{ieee}
\bibliography{fg_geo_arxiv_v2}
}

\end{document}